# Cluster Analysis on Jester Dataset : A Review

Navoneel Chakrabarty
*Machine Learning and Artificial Intelligence Department*
*Liverpool John Moores University*
Liverpool, UK
N.Chakrabarty@2021.ljmu.ac.uk, nc2012@cse.jgec.ac.in

*Abstract*—Unsupervised Machine Learning Paradigms are often the only methodology to rely on, given a Pattern Recognition Task with no target label or annotations being present. In such scenarios, data preparation is a crucial step to be performed so that the Unsupervised Paradigms work with as much perfection as possible. But, when there is no sufficient or missing data being present in each and every instance of a dataset, data preparation becomes a challenge itself. One such case-study is the Jester Dataset that has missing values which are basically ratings given by Joke-Readers to a specified set of 100 jokes. In order to perform a Cluster Analysis on such a dataset, the data preparation step should involve filling the missing ratings with appropriate values followed by cluster analysis using an Unsupervised ML Paradigm. In this study, the most recent and probably the only work that involves Cluster Analysis on the Jester Dataset of Jokes is reviewed and validated with corrections and future scope.

*Index Terms*—Unsupervised Machine Learning, Cluster Analysis, k-Mode Clustering, Restricted Boltzmann Machine (RBM), Jester Dataset

## I. INTRODUCTION

Consumer preferences are of utmost priority for any service-based concern irrespective of the products and services involved. This helps boosting business and consumer satisfaction. Taking this fact under consideration, Cluster Analysis on consumers based on a particular product to understand their tastes across the several kind and genre of that product proves to be instrumental. Such an instance is the Cluster Analysis being applied on the Jester Dataset of Jokes to segment Joke Readers into clusters containing joke readers having similar preferences and tastes. The Jester Dataset consists of 4.1 million anonymous joke-ratings from 73,421 users to 100 specific jokes [1]. Each and every rating is a value between -10.00 to +10.00 with missing ratings denoted by 99. Now, in this paper, the most recent work of Cluster Analysis being applied on the Jester Dataset (by Chakrabarty et al. [2]) is considered for review and further evaluation.

## II. 2 DATA PRE-PROCESSING

Data Pre-processing or Data Preparation plays a crucial role as the Jester Dataset is unclean with a lot of missing values i.e., ratings corresponding to particular (user, joke) pair. To tackle this, Chakrabarty et al. [2] proposed a Restricted Boltzmann Machine (RBM) based Joke Recommender Model prior to applying Cluster Analysis.

Such a recommender model can be utilized to impute the missing values and post that, to carry on the Cluster Analysis on Joke-Readers. But, RBM is a Deep Learning Model, that is trained on binary data as the intuition behind the learning algorithm is based on Bernoulli Distribution. So, the Jester Dataset is transformed such that:

- (User, Joke) Ratings from +7.00 to +10.00 are set to 1 (both inclusive) signifying a 'like'.
- (User, Joke) Ratings from -10.00 to +6.00 are set to 0 (both inclusive) signifying a 'dislike'.
- Missing ratings denoted by 99 are replaced by a new marker '-1'.

Though, this looks convincing but, there are joke-ratings between +6.00 and +7.00 which remains unmentioned. The illustration of the dilemma is shown in Fig 1.

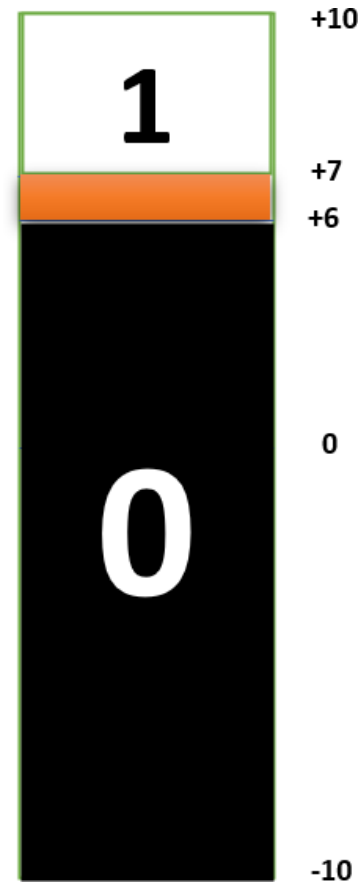

**Fig 1. Here, it seems that the ratings between +6 and +7 are not converted to binary 0/1(section marked by red)**

However, upon checking the implementation, it is confirmed that in point b., (User, Joke) Ratings less than +7.00 are set to 0. The implementation code by Chakrabarty et al. [2] can be found in this link [3]. They also proposed 2 methodologies for Cluster Analysis of Joke-Readers in the Jester Dataset, of which one approach was proven to be feasible. The Proposed Methodologies are discussed in the following section i.e., Section 3.

## III. PROPOSED METHODOLOGY

### A. Methodology - I

The 1st Methodology proposed by Chakrabarty et al. [2] involved the following procedural steps:

1) 80-20 Train-Test Split followed by Training a Restricted Boltzmann Machine (RBM) Model using the Training Set.
2) Performance Analysis of the RBM Model on the Test Set.
3) Using the RBM Model to get the ratings that were not present (missing) in the Jester Dataset and terming it as D2.
4) Application of k-Means Clustering for Cluster Analysis on D2 followed by Cluster Visualization on the same data.

### B. Methodology - II

The 2nd Methodology proposed by Chakrabarty et al. [2] involved the following procedural steps:

1) 80-20 Train-Test Split followed by Training a Restricted Boltzmann Machine (RBM) Model on the Training Set.
2) Performance Analysis of the RBM Model on the Test Set.
3) Using the RBM Model to get the ratings that were not present (missing) in the Jester Dataset and terming it as D2.
4) Obtaining a separate dataset, D1 with only ratings generated by the RBM Model.
5) Application of k-Means Clustering for Cluster Analysis on D1 followed by Cluster Visualization on D2.

Here, both Methodology I and Methodology II are similar until c) i.e., until training, performance analysis of RBM and obtaining missing ratings from it. So, the RBM Model is trained on the Training Set and performance analysis of the same is done on the Test Set in terms of Mean Absolute Error. A Test Mean Absolute Error of 0.1786 is obtained, which is really a decent score. Now, choosing the feasible methodology among Methodology – I and Methodology – II is done on the basis of selection of the optimum number of clusters using Elbow Method based on Within Cluster Sum of Squares (WCSS). So, the Elbow Method is performed by applying k- Means Clustering on D1 and D2 separately and it was found that a perfect elbow was obtained at no. of clusters = 3 on Dataset D1 whereas no elbow was found on Dataset D2. Hence, Methodology – II was followed over Methodology – I with optimum number of clusters to be as 3. After that, the trained k-Means Clustering Model (on Dataset D1) was used to cluster the Joke-Readers with original ratings and imputed ones from the recommender system i.e., Dataset D2.

### C. Cluster Visualization

As the k-Means Clustering was performed on a very high-dimensional data (100 dimensions referring to the 100 jokes), 2-D Visualization or 3-D Visualization with meaningful insights is just not possible, so Chakrabarty et al. [2] formulated a function which they termed as Preference Function defined as,

$$Preference(c_i, j_k) = \frac{Number\ of\ Readers\ belonging\ to\ cluster\ c_i\ and\ liking\ the\ joke\ j_k}{Total\ Number\ of\ Readers\ belonging\ to\ cluster\ c_i}$$

So, the preference values of all the jokes in all the 3 clusters are obtained as 3 vectors of size 100 each and plotted as Line Charts. In this way, the 3 clusters were visualized, showing how varied they are, in terms of the preference values and patterns for all the 100 jokes.

### D. Procedural Imperfection

*1) Cluster-Cluster Overlap:* The joke-reader clusters, so obtained are over-lapping among each other. Although Chakrabarty et al. [2] addressed this, that the similarity in preference values or the preference patterns describe the degree of overlap among the clusters. Moreover, an overlap test is conducted as an important addition in this paper to examine the clusters involved in overlaps. Basically, if data-point(s) (100-dimensional) from Dataset D2 are equidistant from more than one cluster, then the concerned clusters are said to be overlapping. It has been found that clusters 0 2 and clusters 1 2 are overlapping with each other.

*2) Binary Ratings are Categorical:* As the ratings are transformed into binary form as mentioned in Section 2, they are more of categorical (like or dislike) than continuous in nature. As in k-Means, the cluster centroid updation is done by taking mean of all the data-points in a cluster, so the mean of a categorical variable (though in numeric form) does not really make sense. So, the obtained cluster centroid from k-Means Clustering cannot be interpreted directly in order to characterize a cluster.

### E. Corrections to the Proposed Methodology by Chakrabarty et al. [2]

In light of the imperfections addressed above, k-Mode Clustering is applied, which is the most suited for categorical data as it takes the Dissimilarity Metric to assign clusters to data-points and Mode instead of Mean for cluster centroid updation.
Hence, this makes the Cluster Centroids so obtained, meaningful when interpreted. The algorithm of k-Mode Clustering is given in Algorithm 1.

1) From the training data-points, k data-points are selected as k cluster centroids.
2) Assigning all the data-points to the k clusters such that, the cluster centroid having the lowest dissimilarity, $d(x_i, x_j)$ with the concerned data-point, the data-point is assigned that particular cluster.
3) Cluster Centroids are updated by taking the mode of all the data-points inside each of the concerned k clusters.
4) Repeating Steps 2-3, until there is no further change in cluster centroids.

**Algorithm 1. k-Mode Clustering**

*1) Is clustering applicable on Dataset, D1?:* Now, Chakrabarty et al. didn't mention anything about the applicability of clustering algorithms as they missed the applicability test for clustering algorithms. One of those applicability tests is Hopkins Test. So, a Hopkins Statistic is obtained on the dataset, D1 over which clustering is to be applied. The Python's pyclustertend package [4] is used for the Hopkins Test and a Hopkins Statistic so obtained is equal to 0.00019 which is quite closer to 0. Hence, clustering is perfectly applicable on Dataset D1.

*2) Elbow Method to select the optimum number of clusters:* Elbow Method is employed to select the optimum number of clusters in which the Within Cluster Sum of Squares (WCSS) is computed for every number of clusters ranging from 1 to 5.
The Elbow Curve is shown in Fig 2 and a perfect elbow is obtained at number of clusters = 3.

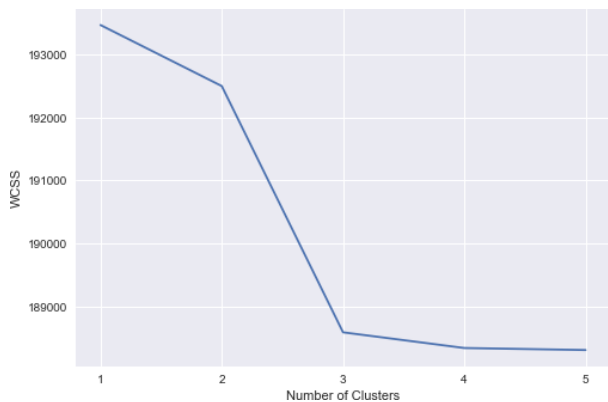

**Fig 2. Elbow Curve on the basis of Within Cluster Sum of Squares (WCSS)**

*3) k-Mode Clustering and Cluster Visualization:* k-Mode Clustering is applied on Dataset, D1 with 3 clusters using Cao initialization. The trained k-Mode Clustering Model is visualized on Dataset D2 that forms the Cluster Visualization by generating Preference Patterns in the form of Line Charts for each and every cluster obeying the Preference Function defined by Chakrabarty et al. [2] and discussed in this paper in Section 3.3.
The preference patterns are shown in Fig 3.

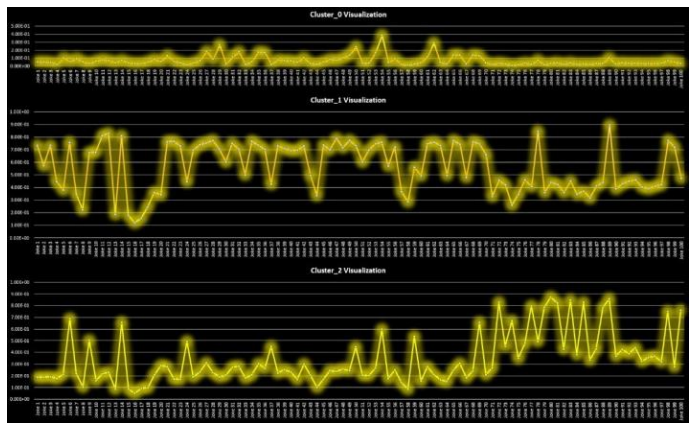

**Fig 3. Cluster Visualizations of the 3 clusters in the form of Preference Patterns as Line Charts**

*4) Cluster Overlap Test:* The Overlap Test is conducted both in terms of the dissimilarity function and WCSS. It has been found that the overlap still exists but has been reduced by half as previously there has been overlapping between clusters 0_2 and 1_2. But in this approach following k-Mode, there is an overlapping found only between the users of clusters 0_2. So, even though the overlapping issue still persists, it has been reduced to a great extent.

IV. CONCLUSION

So, the approach of k-Mode Clustering mitigated the procedural imperfections in the paper by Chakrabarty et al. [2] to an extent. The cluster overlap issue has been overcome by 50% and k-Mode Clustering works on the dissimilarity function with cluster centroid updation being done on the basis of mode unlike k-Means Clustering in which mean is used for cluster centroid updation.
Hence, the cluster centroids, so obtained from k-Mode Clustering are meaningful and interpretable for categorical data (as ratings are binary). The future scope of this work lies in applying or proposing methodologies for cluster analysis to rectify the issue of Cluster Overlap completely.

REFERENCES

[1] Jester Dataset, https://goldberg.berkeley.edu/jester-data/
[2] Chakrabarty N, Rana S, Chowdhury S, Maitra R. RBM based joke recommendation system and joke reader segmentation. InInternational Conference on Pattern Recognition and Machine Intelligence 2019 Dec 17 (pp. 229-239). Springer, Cham.
[3] https://github.com/navoneel1092283/Joke_RecommendationSegmentation
[4] https://pypi.org/project/pyclustertend/